\documentclass[5p, twocolumn]{elsarticle}

\usepackage{multirow} 
\usepackage{amsmath,amssymb,amsfonts}
\usepackage{graphicx}

\usepackage{algorithmic}
\usepackage{textcomp}
\usepackage{xcolor}
\usepackage{subfigure}
\usepackage{multirow}
\usepackage{booktabs}
\usepackage{soul}
\usepackage{hyperref}

\usepackage{lineno}

\newsavebox\CBox
\def\textBF#1{\sbox\CBox{#1}\resizebox{\wd\CBox}{\ht\CBox}{\textbf{#1}}}

\journal{Elsevier}

\begin{document}

\begin{frontmatter}

\title{CEKD:Cross Ensemble Knowledge Distillation for Augmented Fine-grained Data}

\author[mymainaddress]{Ke Zhang}
\ead{ke.zhang@hdu.edu.cn}

\author[mymainaddress]{Jin Fan\corref{mycorrespondingauthor}}
\cortext[mycorrespondingauthor]{Corresponding author}
\ead{fanjin@hdu.edu.cn}

\author[mysecondaryaddress]{Shaoli Huang}
\ead{shaoli.huang@sydney.edu.au}

\author[mythirdaddress]{Yongliang Qiao}
\ead{yongliang.qiao@sydney.edu.cn}

\author[mymainaddress]{Xiaofeng Yu}
\ead{xiaofengyu@hdu.edu.cn}

\author[mymainaddress]{Feiwei Qin}
\ead{qinfeiwei@hdu.edu.cn}

\address[mymainaddress]{College of Computer Science and Technology, Hangzhou Dianzi University, Hangzhou, China}
\address[mysecondaryaddress]{UBTECH Sydney AI Centre, School of Computer Science, FEIT, University of Sydney, Sydney, Australia}
\address[mythirdaddress]{Australian Centre for Field Robotics (ACFR), Faculty of Engineering, University of Sydney, Sydney, Australia}

\begin{abstract}
Data augmentation has been proved effective in training deep models. Existing data augmentation methods tackle fine-grained problem by blending image pairs and fusing corresponding labels according to the statistics of mixed pixels, which produces additional noise harmful to the performance of networks. Motivated by this, we present a simple yet effective cross ensemble knowledge distillation (CEKD) model for fine-grained feature learning. We innovatively propose a cross distillation module to provide additional supervision to alleviate the noise problem, and propose a collaborative ensemble module to overcome the target conflict problem.
The proposed model can be trained in an end-to-end manner, and only requires image-level label supervision. Extensive experiments on widely used fine-grained benchmarks demonstrate the effectiveness of our proposed model. Specifically, with the backbone of ResNet-101, CEKD obtains the accuracy of 89.59\%, 95.96\% and 94.56\% in three datasets respectively, outperforming state-of-the-art API-Net  by 0.99\%, 1.06\% and 1.16\%.

\end{abstract}


\end{frontmatter}


\section{Introduction}
Fine-grained recognition aims to distinguish subcategories belonging to the same category, e.g., bird species~\cite{CUB}, cars~\cite{Cars} and aircrafts~\cite{Aircraft}, which persists to be a challenging problem due to large intra-class variance and small inter-class variance. To this end, a number of fine-grained frameworks have been designed to find the discriminative semantic parts~\cite{huang2016part,hu2018squeeze,zhang2019learning,du2020fine}. However, these frameworks more or less exist overfitting problem. To alleviate overfitting problem, deep neural network with data augmentation method has made enormous advances. Among them, mixed-based method such as MixUp~\cite{mixup}, CutMix~\cite{yun2019cutmix} and SnapMix~\cite{huang2020snapmix} have drawn increasing attention. Although these data augmentation methods play a significant role in avoiding overfitting, they produce additional noise during the mixing operations result in  unreliable of data augmentation~\cite{gong2021keepaugment}. In addition, it's well known that knowledge distillation~\cite{hinton2015distilling} is a widely used model compression method. However, offline distillation has many defects(e.g., a large-scale model of high performance is not always available), while online distillation enable teachers and students to update end-to-end with effective parallel computing simultaneously.\par
  
Inspired by the above observation, we provide a novel insight to combine data augmentation methods with online knowledge distillation technique, and propose a cross ensemble knowledge distillation (CEKD) model. To overcome the noise interference, we propose a cross distillation module to reconstruct the input path of the network to obtain additional supervision. The main idea of cross distillation is that both teacher and student network have expertise in their respective fields, that is, they are good at handling different augmentation methods. If the input of one is regarded as noise to the other, the anti-interference ability for data noise of both sides can be improved.
\par
In the field of knowledge distillation, target conflict problem is commonly caused by the performance gap between teacher and student network. It is known, a teacher with high quality is essential for optimizing a good student. It is confirmed that if the teacher is not well optimized and provides noisy supervision, the risk that soft target and ground truth conflict with each other becomes high~\cite{KDCL}.

Due to the target conflict problem, we propose a collaborative ensemble module to dynamically ensemble and optimize all final output logits. It is worth noting that our ensemble method is not only limited to integrating and optimizing the outputs of knowledge distillation, but also to cooperating with cross distillation (CD) module and optimizing the output of CD simultaneously. Through dynamic integration of outputs, the performance gap between teacher and student network can be better compensated, which is beneficial for improving the performance of teacher network. Finally, combining the two modules, the proposed methods can comprehensively capture discriminative regions and improve the robustness of our model.\par

Overall, the main contributions of our work are summarized as follows:
\begin{itemize}
    \item 
    We provide a novel insight to combine data augmentation methods with knowledge distillation technique, and propose cross ensemble knowledge distillation (CEKD) model.
    \item 
    A cross distillation module is proposed to enhance the robustness of the model against noise, meanwhile a collaborative ensemble module is proposed  to alleviate the target conflict problem.
    \item 
    Experiments  on three challenging fine-grained datasets (CUB-200-2011~\cite{CUB}, Stanford-Cars~\cite{Cars}, and FGVC-Aircraft~\cite{Aircraft}) illustrates that our model outperforms other state-of-the-art methods.
\end{itemize}

The rest of the paper is organised as follows. Section~\ref{related works}
describes the related works, and Section~\ref{preliminary} introduces the preliminary. Section~\ref{approach} illustrates the proposed method in detail. Section~\ref{experiments} and Section~\ref{further experiments} present the implementations and experimental results, followed by the conclusion in Section~\ref{conclution}.

\section{Related Works}
\label{related works}
\subsection{Fine-grained recognition}
The field of fine-grained recognition~\cite{wang2020multi, wang2020kernelized, wang2020attribute, wang2021discriminative} has been widely studied. The early part-based method used additional strong supervision information to identify object categories, such as part annotations and bounding box annotations. Since manual part annotations consumed too many resources, some weakly supervised localization methods using only image-level annotations were proposed. ~\cite{oquab2015object} first assumed the possibility of an object's location using max-pooling layer and modified the loss function only from the image-level label. ~\cite{CAM} proved that the features extracted by CNN contain location information through CAM, and this location information can be transferred to other cognitive tasks. ~\cite{Grad-CAM} improved CAM and proposed Grad-CAM to make CNN-based model more transparent by generating visual interpretation. ~\cite{zhang2018adversarial} automatically located the integrated region of interest through a weakly supervised way, and directly selected the class-specific feature map from the final convolution layer. ~\cite{zhuang2020learning} adopts an attentive pairwise interaction network to identify differences by comparing image pairs. ~\cite{gao2020channel} models channel interaction to mine semantically complementary information. Both of them require a large number of complex submodules to model pairwise or channel-wise interactions, which is extremely time-consuming. In contrast, we tackle this problem using lightweight method only with single backbone to improve efficiency. Different from the above methods, we innovatively combining data augmentation with the knowledge distillation technique to reduce annotation resource consumption. More specifically, we propose cross distillation and collaborative ensemble to enhance the robustness of our model against noise and alleviate target conflict problems.

\subsection{Data augmentation}
Data augmentation is one effective way to reduce the over-fitting problem in the process of DNN training~\cite{simard1998transformation}. At present, the most advanced methods can be divided into two categories: regional elimination method and data mixing method. The former prevented CNN over-fitting by removing part of the image patch, which encouraged the network to use the information of the entire image, rather than relying on a small part of the specific visual features. Cutout~\cite{cutout} is a typical representative method to enhance data by removing the square area. The latter is data mixing method, which has received more attention than the former. It mixed two sets of images at pixel level, and combined the label information of the image. Among these methods, ~\cite{mixup} first proposed mixup which randomly selected two samples and their labels for simple random weighted summation. ~\cite{summers2019improved} improved the mixup method by using a more generalized nonlinear combination method. ~\cite{yun2019cutmix} proposed cutmix which shows better performance by removing part of the image and filling it with the same size part of another image and determining the label weight according to the pixel ratio. In the latest research, snapmix~\cite{huang2020snapmix} developed the cutmix method by using CAM to obtain semantic-relatedness proportion to mix labels more consistently with the mixed image’s semantic structure. However, these data augmentation methods generate additional noises during the mixing operations, which makes them become unreliable. In our work, three representative data mixing methods, namely, mixup, cutmix and snapmix are selected for comparative experiments.

\subsection{Knowledge distillation}
The goal of knowledge distillation is to train a simple and effective student network, which learns from hard labels and soft labels provided by teacher network~\cite{ba2013deep,hinton2015distilling}. According to whether the teacher network is trained with the student network, knowledge distillation can be divided into two categories: offline distillation~\cite{passalis2018learning,heo2019knowledge,mirzadeh2020improved,li2020few} and online distillation~\cite{xie2019training,chen2020online,chung2020feature,KDCL}. Most of the current methods were using offline distillation for two-stage training, first training the teacher network, and then the trained teacher network is applied to guide the training of student network. Although offline distillation is simple and efficient, there are still many limitations, especially when a high-performance teacher model is not available. Online distillation is proposed to overcome the limitation of offline distillation, in which teacher and student networks were trained end-to-end together. Our model is similar to online distillation in jointly training both teacher and student networks. Differently, we design an extra cross distillation module to reconstruct the input path of the networks for obtaining additional supervision, enabling the model to be trained more stable against data noise.

\begin{figure*}[ht] 
	\centering 
	\includegraphics[width=\linewidth]{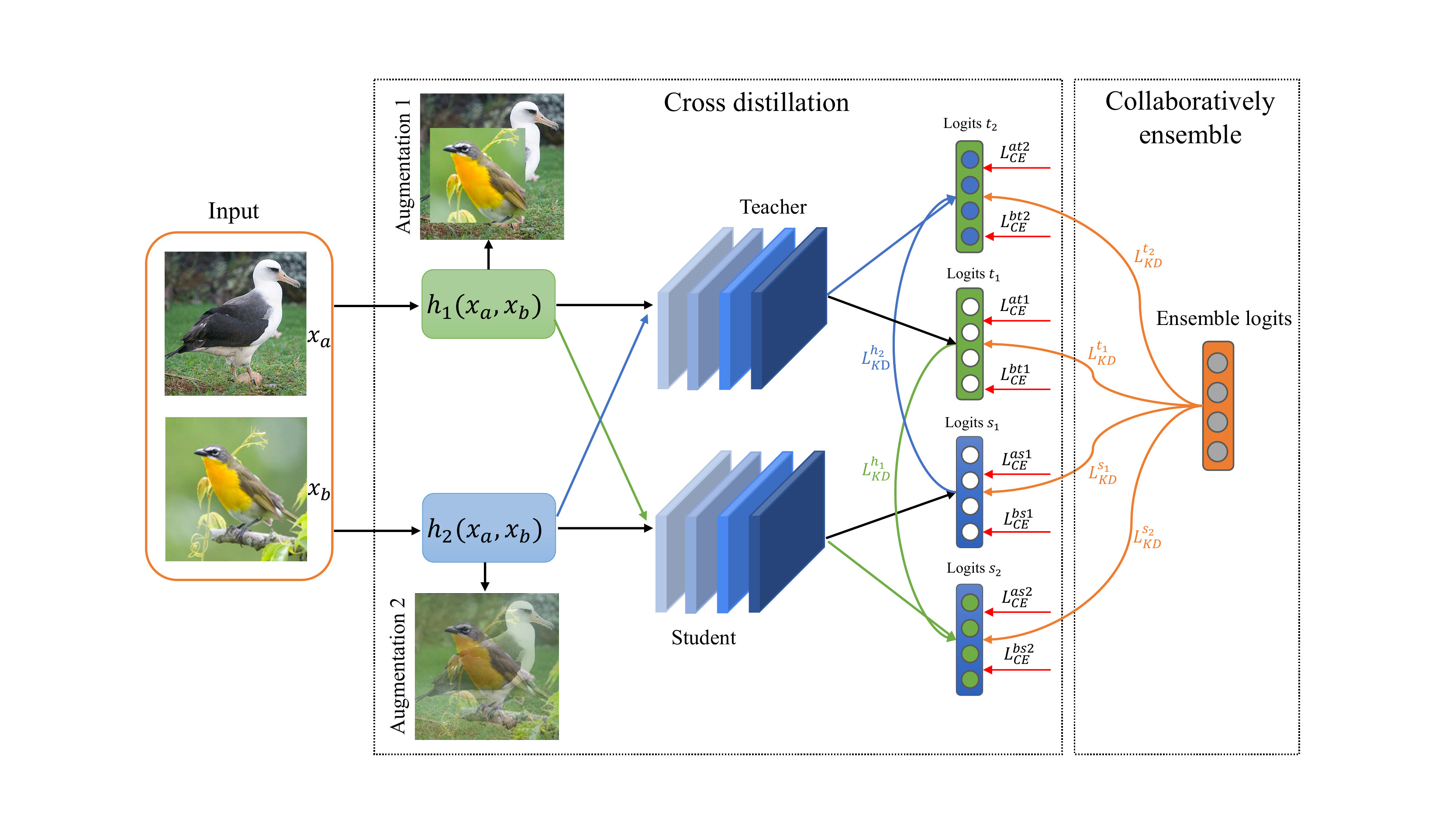} 
	\caption{Overview of the CEKD architecture. $x_a$ and $x_b$ are two input images from different categories, and they are augmented by data mixing methods $h_1\left( \cdot \right)$ and $h_2\left( \cdot \right)$ respectively to get $h_1\left( x_a,x_b \right)$ and $h_2\left( x_a,x_b \right)$. Black lines represent for the input path of normal knowledge distillation framework. Green and blue lines represent for cross input path we proposed. Red lines denote the cross-entropy loss ($L_{CE}$) with hard targets. For brevity, we omit the mixing ratio for $L_{CE}$ within one of the output logits.} 
	\label{fig:architecture} 
\end{figure*}

\section{Preliminary}
\label{preliminary}
In this section, we introduce some background by formulations on data augmentation and knowledge distillation, which are used in our model. We start by briefly illustrating  three data mixing methods in Section~\ref{data mixing methods}, then introducing general knowledge distillation in Section~\ref{knowledge distillation}.
\subsection{Data mixing methods}
\label{data mixing methods}
Data mixing methods first mix the images, and then generate the coefficients to mix the corresponding labels. Note that the original training dataset $\left \{ \left ( x_i,y_i\right )\mid i\in\left [ 0,1,...,N-1 \right ]   \right \} $, where $x_i \in R^{3\times H \times W}$, and $y_i$ correspond to original images and labels, respectively. Given a data pair $\left ( x_a,y_a \right )$ and $\left ( x_b,y_b \right )$, with a random value $\lambda$ sampled from a beta distribution $Beta\left ( \alpha,\alpha \right)$, it generates a new image $I$ and two label weights $w_a$ and $w_b$ according to $\lambda$. Here, $\alpha$ is a hyperparameter. $w_a$ and $w_b$ are corresponding to label $y_a$ and $y_b$, respectively.
The most advanced and representative data mixing methods are MixUp~\cite{mixup}, CutMix~\cite{yun2019cutmix}, and SnapMix~\cite{huang2020snapmix}.  \par
$MixUp$ uses linear combination to mix images and labels, which can be expressed as:
\begin{equation}
    \begin{split}
        &I= \lambda \times x_a + \left( 1-\lambda \right) \times x_b, \\
        &w_a=\lambda, w_b=1-\lambda,
    \end{split}
\end{equation}

$CutMix$ adopts cut-and-paste operation on images and mix labels according to the ratio of cutting area:
\begin{equation}
    \begin{split}
        &I= \left( 1-M_\lambda \right) \odot x_a + M_\lambda \odot x_b, \\
        &w_a=1-\lambda, w_b=\lambda,
    \end{split}
\end{equation}
where,  $\odot$ denotes element-wise multiplication and $M_\lambda$ is a binary mask of a random box region with area ratio $\lambda$.\par
$SnapMix$ use CAM to perform semantic analysis on the images further on the basis of CutMix, which performs similar cut-and paste operation. The semantic ratios of the two regions are calculated respectively, and the labels are mixed according to the semantic ratios.  \par
\begin{equation}
    \begin{split}
        &I= \left( 1-M_{\lambda^a} \right) \odot x_a + T_\theta\left( M_{\lambda^b} \odot x_b \right),\\
        &S\left( x_i \right) = \frac{CAM\left( x_i \right)}{sum\left( CAM\left( x_i \right) \right)},\\
        &w_a=1-sum\left( M_{\lambda^a} \odot S\left( x_a \right) \right), \\
        &w_b=sum\left( M_{\lambda^b} \odot S\left( x_b \right) \right),
    \end{split}
\end{equation}
where, $M_{\lambda^a}$ and $M_{\lambda^b}$ are two binary mask of random box regions with area ratio $\lambda^a$ and $\lambda^b$, $T_\theta$ is the transformation function to make the cutting region of $x_b$ match the cutting region of $x_a$. $S\left( x_i \right)$ denotes the function to calculate semantic percent map of an image $x_i$ by normalizing the CAM to sum to one. \par

\subsection{Knowledge distillation}
\label{knowledge distillation}
Knowledge distillation is to optimize student network under the guidance of teacher network. The loss function uses the soften output of the teacher network and the student network to calculate the KL divergence. The formulation is as follows:
\begin{equation}
    L_{KD}=\frac{1}{n}\sum_{n}^{i=1}T^2KL\left( \mathbf{p}_i,\mathbf{q}_i \right),
\end{equation}
where, $n$ is the batch size, $T$ is the temperature, $\mathbf{p}$ and $\mathbf{q}$ represent for the soften probability distribution given by teacher and student network. Note the output of teacher and student network as $logits\ t$ and $logits\ s$. Then the soft target $\mathbf{p}=softmax\left( t /T\right)$ and $\mathbf{q}=softmax\left( s /T\right)$.

\section{Approach}
\label{approach}
As illustrated in Fig.~\ref{fig:architecture}, the proposed cross ensemble knowledge distillation (CEKD) model dynamically generate soft targets in an online manner.
The entire model is composed of teacher and student branches, where the teacher and student network adopt the same backbone network. The details of experimental settings will be introduced in Section~\ref{Implementation details}. We take the same way to prepocess images as SnapMix~\cite{huang2020snapmix}. After the process of different data augmentation operations $h_1\left( \cdot \right)$ and $h_2\left( \cdot \right)$, the mixed images are fed into teacher and student backbone, and get the output $logits\ t_1$ and $logits\ s_1$. Red lines denote the cross-entropy loss ($L_{CE}$) with hard targets. In order to reduce the noise and target conflicting, we propose cross distillation and collaborative ensemble methods to enhance the robustness of our knowledge distillation model.
 We start by introducing cross distillation method in Section~\ref{cross distillation}, then collaborative ensemble method in Section~\ref{collaborative ensemble}.
\subsection{Cross distillation}
\label{cross distillation}
Many works have demonstrated that data augmentation can improve the generalization ability of the model and prevent over-fitting~\cite{mixup,yun2019cutmix,huang2020snapmix}. Nonetheless, it is still suffer from the problem of introducing a large amount of data noise~\cite{gong2021keepaugment}. Hence, we propose cross distillation to enhance the robustness of our model against noise. Contrary to the way we input the mixed image before, we cross input the augmented image $h_1\left( x_a,x_b \right)$ and $h_2\left( x_a,x_b \right)$ into student and teacher network to obtain $logits\ t_2$ and $logits\ s_2$. As a matter of fact, for the teacher network, it always receives the input being augmented by $h_1$ operation. When the teacher network which is good at processing $h_1\left( x_a,x_b \right)$ obtains the input processed by $h_2$, $h_2\left( x_a,x_b \right)$ is actually noise for the teacher network. Notably, if the used student network  is good at processing $h_2\left( x_a,x_b \right)$ to guide the teacher network to deal with noise interference, the robustness of the teacher network can be apparently improved. Likewise, $h_1\left( x_a,x_b \right)$ can also be seen as noise for student networks which is expert in dealing with $h_2\left( x_a,x_b \right)$, we can use teacher network to guide student network to cope with noise interference. More specifically, we use KD loss to instruct each other as
\begin{equation}
    \begin{split}
        &L_{KD}^{h_1}=\frac{1}{n}\sum_{n}^{i=1}T^2KL\left( \sigma \left ( TE\left ( h_1\left ( x_a,x_b \right )  \right )  \right ), \sigma \left ( ST\left ( h_1\left ( x_a,x_b \right )  \right )  \right ) \right),\\
        &L_{KD}^{h_2}=\frac{1}{n}\sum_{n}^{i=1}T^2KL\left( \sigma \left ( ST\left ( h_2\left ( x_a,x_b \right )  \right )  \right ), \sigma \left ( TE\left ( h_2\left ( x_a,x_b \right )  \right )  \right ) \right),
    \end{split}
\end{equation}
where $TE\left( \cdot \right)$ and $ST\left( \cdot \right)$ represent for teacher and student network respectively, $\sigma\left( \cdot \right)$ is softmax function.


\subsection{Collaborative ensemble}
\label{collaborative ensemble}
Due to the target conflict problem, we propose an optimized collaborative ensemble method to alleviate this problem by ensembling output logits dynamically.\par
Firstly, we concatenate the four output logits by column into matrix $M =\left[ t^T_1;t^T_2;s^T_1;s^T_2 \right] \in R^{c\times 4}$, where $c$ is the total number of categories. For arbitrary logits $l$, in which each value implies the corresponding class probability information. Its probability distribution $\mathbf{p}$ can be expressed by softmax function 
\begin{equation}
    \mathbf{p} = softmax(l), 
\end{equation}

Assuming that the ground truth category of logits $l$ is $c$, we consider the logits corresponding to other classes except the ground truth category as $l^c$. It is worth noting that if $l^c$ becomes smaller, the probability of mis-classification will reduce, that is, the cross entropy loss will reduce. Thus, a neat way to generate the corresponding row value of ensemble logits is to  select the minimum value of each row of matrix $M$. However, considering that our model has two image mixing methods, there will be two synthetic true labels. Consequently, logits in matrix $M$ need to be split into two groups. We split the matrix $M$ into matrices $E^1 =\left[ t^T_1;s^T_2 \right]\in R^{c\times2}$ and $E^2 =\left[ s^T_1;t^T_2 \right]\in R^{c\times2}$ according to its corresponding data mixing method. Thus the corresponding ensemble logits can be gotten, denoted as $e_1,e_2$.  
\begin{equation}
    e_{t,j}=min\left\{E^t_{j,i}\mid i=1,2 \right\},
\end{equation}
where $t\in \left\{ 1,2 \right\}$ denotes two serial numbers of ensemble logits, $e_{t,j}$ is the element of the j-th row of the t-th ensemble logits, $E^t_{j,i}$ is the element of the j-th row and the i-th column in $E^t$.\par
Therefore, we can obtain four KD losses by calculating the KL divergence between four output logits and ensemble logits $e_t,t\in\left\{ 1,2 \right\}$. 
\begin{equation}
    \begin{split}
        &L_{KD}^{t_1}=\frac{1}{n}\sum_{n}^{i=1}T^2KL\left( \sigma \left ( e_1 \right ), \sigma \left ( t_1 \right) \right),\\
        &L_{KD}^{t_2}=\frac{1}{n}\sum_{n}^{i=1}T^2KL\left( \sigma \left ( e_2 \right ), \sigma \left ( t_2 \right) \right),\\
        &L_{KD}^{s_1}=\frac{1}{n}\sum_{n}^{i=1}T^2KL\left( \sigma \left ( e_2 \right ), \sigma \left ( s_1 \right) \right),\\
        &L_{KD}^{s_2}=\frac{1}{n}\sum_{n}^{i=1}T^2KL\left( \sigma \left ( e_1 \right ), \sigma \left ( s_2 \right) \right),
    \end{split}
\end{equation}
The final KD loss $\mathcal{L}_{KD}$ is a combination of six losses from the former two modules,
\begin{equation}
\label{all loss}
    \mathcal{L}_{KD} = \lambda_{1}L_{KD}^{h_1}+\lambda_{2}L_{KD}^{h_2}+\lambda_{3}L_{KD}^{t_1}+\lambda_{4}L_{KD}^{t_2}+\lambda_{5}L_{KD}^{s_1}+\lambda_{6}L_{KD}^{s_2},
\end{equation}
where $\lambda_{1...6}\in \mathbb{R}^+$ are the coefficients to balance the contribution from each loss.

\begin{table*}[ht]
\caption{Performance comparison(Acc.\%) with original data mixing methods on CUB, Cars and Aircraft using backbones ResNet-18 and ResNet-34. All methods' improvement with CEKD compared to their original performance are shown in the brackets.}
\centering
\renewcommand\arraystretch{1.5}
\renewcommand\tabcolsep{6.0pt} 
\begin{tabular}{cccccccc}
\toprule
        &         & \multicolumn{2}{c|}{CUB}                          & \multicolumn{2}{c|}{Cars}                          & \multicolumn{2}{c}{Aircraft} \\ 
        &         & Res18        & \multicolumn{1}{c|}{Res34}        & Res18        & \multicolumn{1}{c|}{Res34}        & Res18         & Res34        \\ \hline
        
Original & MixUp & 83.17 & \multicolumn{1}{c|}{85.22} & 91.57 & \multicolumn{1}{c|}{93.28} & 89.82  & 91.02 \\
Original & CutMix  & 80.16 & \multicolumn{1}{c|}{85.69} & 92.65 & \multicolumn{1}{c|}{93.61} & 89.44  & 91.26 \\ 
Original & SnapMix & 84.29 & \multicolumn{1}{c|}{87.06} & 93.12 &
\multicolumn{1}{c|}{93.95} & 90.17  & 92.36 \\  \hline

Teacher & SnapMix & 84.34(+0.05) & \multicolumn{1}{c|}{87.90(+0.84)} & 93.16(+0.04) & \multicolumn{1}{c|}{94.53(+0.58)} & 91.02(+0.85)  & 93.17(+0.81) \\
Student & CutMix  & 82.17(+2.01) & \multicolumn{1}{c|}{86.40(+0.71)} & 93.01(+0.44) & \multicolumn{1}{c|}{93.96(+0.35)} & 90.08(+0.64)  & 92.10(+0.84) \\ \hline
Teacher & SnapMix & \textBF{85.22(+0.93)} & \multicolumn{1}{c|}{\textBF{88.32(+1.26)}} & \textBF{93.87(+0.75)} & \multicolumn{1}{c|}{\textBF{94.61(+0.66)}} & \textBF{91.29(+1.12)}  & \textBF{93.65(+1.29)} \\
Student & MixUp   & 84.45(+1.28) & \multicolumn{1}{c|}{86.46(+1.24)} & 92.63(+1.06) & \multicolumn{1}{c|}{93.50(+0.22)} & 90.64(+0.82)  & 92.48(+1.46) \\ \bottomrule
\end{tabular}
\label{tab:performance on res18 and res34}
\end{table*}

\begin{table*}[ht]
\caption{Performance comparison(Acc.\%) with original data mixing methods on CUB, Cars and Aircraft using backbones ResNet-50 and ResNet-101. All methods' improvement with CEKD compared to their original performance are shown in the brackets.}
\centering
\renewcommand\arraystretch{1.5}
\renewcommand\tabcolsep{6.0pt} 
\begin{tabular}{cccccccc}
\toprule
        &         & \multicolumn{2}{c|}{CUB}                          & \multicolumn{2}{c|}{Cars}                          & \multicolumn{2}{c}{Aircraft} \\ 
        &         & Res50        & \multicolumn{1}{c|}{Res101}       & Res50        & \multicolumn{1}{c|}{Res101}       & Res50         & Res101       \\ \hline
        
Original & MixUp & 86.23 & \multicolumn{1}{c|}{87.72} & 93.96 & \multicolumn{1}{c|}{94.22} & 92.24  & 92.89 \\
Original & CutMix  & 86.15 & \multicolumn{1}{c|}{87.92} & 94.18 & \multicolumn{1}{c|}{94.27} & 92.23  & 92.29 \\ 
Original & SnapMix & 87.75 & \multicolumn{1}{c|}{88.45} & 94.30 &
\multicolumn{1}{c|}{94.44} & 92.80  & 93.74 \\  \hline        
        
Teacher & SnapMix & 87.93(+0.18) & \multicolumn{1}{c|}{89.38(+0.93)} & 95.00(+0.70) & \multicolumn{1}{c|}{95.65(+1.21)} & 93.24(+0.44)  & 93.97(+0.23) \\
Student & CutMix  & 86.37(+0.22) & \multicolumn{1}{c|}{88.22(+0.30)} & 94.82(+0.64) & \multicolumn{1}{c|}{95.34(+1.07)} & 92.47(+0.24)  & 92.59(+0.30) \\ \hline
Teacher & SnapMix & \textBF{88.41(+0.66)} & \multicolumn{1}{c|}{\textBF{89.59(+1.14)}} & \textBF{95.65(+1.35)} & \multicolumn{1}{c|}{\textBF{95.96(+1.52)}} & \textBF{93.48(+0.68)}  & \textBF{94.56(+0.82)} \\
Student & MixUp   & 86.60(+0.37) & \multicolumn{1}{c|}{88.24(+0.52)} & 94.99(+1.03) & \multicolumn{1}{c|}{95.46(+1.24)} & 93.02(+0.78)  & 93.46(+0.57) \\ \bottomrule
\end{tabular}
\label{tab:performance on res50 and res101}
\end{table*}

\section{Experiments}
\label{experiments}
In this section, in order to verify the effectiveness of our method, extensive experiments on three fine-grained datasets are conducted. We start by introducing the experimental datasets, and then experimental settings, followed by performance comparison with original data mixing methods and state-of-the-art methods (e.g.,  LIO~\cite{zhou2020look}, CIN~\cite{gao2020channel} and API-Net~\cite{zhuang2020learning}). 
\subsection{Datasets}
The empirical evaluation is performed on three widely used fine-grained benchmarks: Caltech-USCD Birds(CUB-200-2011)~\cite{CUB}, FGVC-Aircraft~\cite{Aircraft}, and Standford-Cars~\cite{Cars}. Note that we only use the image-level labels in our all experiments.\par
\textbf{CUB-200-2011} is the most widely used dataset with 200 species of wild birds which contains 11788 images.  The training set and the testing set contain 5994 and 5794 instances respectively with 30 images per species for training.\par
\textbf{FGVC-Aircraft} dataset contains 10,000 images of 100 categories, among which 6,667 used for training and 3,333 used for testing. Each class contains 100 aircraft images.\par
\textbf{Stanford-Cars} dataset consists of 16185 images from 196 categories which are split into 8144 training images and 8041 test images.

\subsection{Implementation details}
\label{Implementation details}
In our implementation, we evaluate our method on a series of backbone networks (ResNet-18,34,50,101~\cite{resnet}). We adopt the results reported by the corresponding data augmentation method as baseline. Since SnapMix has the best performance compared to MixUp and CutMix, it is adopted as the teacher network in our experiment setting while CutMix or MixUp as student network. In addition, each baseline network is pre-trained on ImageNet and finetuned on its target dataset. During the network training, the images are first resized into 512×512, then randomly cropped into 448$\times$448 with horizontally flipped. We train the models for 200 epochs using SGD with the momentum of 0.9 and the learning rate decays by factor 0.1 every 80 epoch. During inference, the original images are resized and center cropped to 448×448. We set the probability of being augmented to 0.5 for MixUp and 1.0 for CutMix and SnapMix. 

\begin{table}[ht]
\caption{Performance comparison(Acc.\%) with state-of-the-art methods on CUB, Cars and Aircraft.}
\centering
\renewcommand\arraystretch{1.5}
\renewcommand\tabcolsep{4.0pt} 
\begin{tabular}{ccccc}
\toprule
\multirow{2}{*}{Method} & \multirow{2}{*}{Backbone} & \multicolumn{3}{c}{Accuracy(\%)}                 \\ \cline{3-5} 
                        &                           & CUB            & Cars           & Aircraft       \\ \hline
DeepLAC\cite{lin2015deep}                 & VGG                       & 80.3           & -              & -              \\
Part-RCNN\cite{zhang2014part}               & VGG                       & 81.6           & -              & -              \\
RA-CNN\cite{fu2017look}                  & VGG                       & 85.3           & 92.5           & 88.1           \\
MA-CNN\cite{zheng2017learning}                  & VGG                       & 86.5           & 92.8           & 89.9           \\
MAMC\cite{sun2018multi}                    & Res-50                    & 86.2           & 92.8           & -              \\
NTS\cite{yang2018learning}                     & Res-50                    & 87.5           & 93.3           & 91.4           \\
API-Net\cite{zhuang2020learning}                 & Res-50                    & 87.7           & 94.8           & 93.0           \\
Cross-X\cite{luo2019cross}                 & Res-50                    & 87.7           & 94.5           & 92.6           \\
DCL\cite{chen2019destruction}                     & Res-50                    & 87.8           & 94.5           & 93.0           \\
DTB-Net\cite{zheng2019learning}                 & Res-50                    & 87.5           & 94.1           & 91.2           \\
CIN\cite{gao2020channel}                     & Res-50                    & 87.5           & 94.1           & 92.6           \\
LIO\cite{zhou2020look}                     & Res-50                    & 88.0           & 94.5           & 92.7           \\
MAMC\cite{sun2018multi}                    & Res-101                   & 86.5           & 93.0           & -              \\
DTB-Net\cite{zheng2019learning}                 & Res-101                   & 88.1           & 94.5           & 91.6           \\
CIN\cite{gao2020channel}                     & Res-101                   & 88.1           & 94.5           & 92.8           \\
API-Net\cite{zhuang2020learning}                 & Res-101                   & 88.6           & 94.9           & 93.4           \\ \hline
Snapmix                & Res-50                    & 87.75          & 94.3           & 92.8           \\
Snapmix + \textBF{CEKD}         & Res-50                    & \textBF{88.41} & \textBF{95.65} & \textBF{93.48} \\
Snapmix                & Res-101                   & 88.45          & 94.44          & 93.74          \\
Snapmix + \textBF{CEKD}         & Res-101                   & \textBF{89.59} & \textBF{95.96} & \textBF{94.56} \\ \bottomrule  
\end{tabular}
\label{tab:performance with state-of-the-art}
\end{table}

\subsection{Comparison with original data mixing methods}
For brevity, we use short names CUB, Cars and Aircraft to denote corresponding datasets Caltech-USCD Birds-200-2011, Standford Cars, and FGVC-Aircraft. Table.~\ref{tab:performance on res18 and res34} and Table.~\ref{tab:performance on res50 and res101} presents  comparison results on three datasets.  Here, we  use SnapMix as teacher network, with the combination of CutMix or MixUp as student network, and adopt the results reported by the corresponding data augmentation method as baseline. In brackets, we further record the improvement of each original data mixing method compared to it being processed by our proposed CEKD. First, it can be observed that both teacher and student networks have been significantly improved. Compared with original SnapMix in CUB dataset, teacher network in our method obtains a better result with an improvement of 1.26\% and 1.14\% on the backbone of Res-34 and Res-101, respectively. Meanwhile, the corresponding student network in our method reach an improvement of 1.24\% and 0.52\% compared to original MixUp. The experimental results show that our proposed CEKD method is effective whether it is the combination of CutMix and SnapMix or the combination of MixUp and SnapMix. Moreover, we can find that when MixUp is adopted as student network, the improvement of teacher network (SnapMix) is much higher than that when CutMix is used as a student network. This might mainly because CutMix and SnapMix use cut-and-paste operation to make the model pay more attention to the local criminative region while MixUp focuses on the global feature. Although MixUp belongs to mixed-based method as the former two methods, it tends to retain the complete global structure of the image while mixing. Therefore, MixUp and SnapMix can complement each other to improve the final performance. In addition, the proposed method can still improve the teacher network in the CUB dataset which needs more subtle identification between categories. However, due to the reason that CUB dataset is more sensitive to label noise, the improvement on CUB is lower than the other two datasets.


\subsection{Comparison with state-of-the-art methods}
In this section, the proposed CEKD method is compared with other state-of-the-art methods on three fine-grained benchmark datasets. Notably, we adopt the performance of snapmix as baseline. In Table.~\ref{tab:performance with state-of-the-art}, it can be observed that our method achieve better performance than  the state-of-the-art methods by a considerable margin. More precisely, with the backbone of ReNet-101, our method achieve the accuracy of 89.59\%, 95.96\% and 94.56\% on CUB, Cars and Aircraft respectively, which outperforms API-Net by 0.99\%, 1.06\% and 1.16\%. Moreover, API-Net and CIN are two of the state-of-the-art models. API-Net adopts an attentive pairwise interaction network to identify differences by comparing image pairs while CIN models channel interaction to mine semantically complementary information. Both of them require a large number of complex submodules to model pairwise or channel-wise interactions, which is extremely time-consuming. In contrast, our lightweight method with single backbone not only strengthens the baseline, but also outperforms advanced and intricate CIN and API-Net. It is proved that our proposed efficient CEKD model can be used in various fine-grained recognition tasks to obtain better recognition accuracy.

\section{Further experiments}
\label{further experiments}

\subsection{Performance of using other backbones}
So far we have done experiments using ResNet as backbone. In order to prove the effectiveness of our proposed method more comprehensively, we have also made comparisons on other backbones. As shown in Table.~\ref{tab:using other backbones}, we verify the performance of our method on two other backbones that is InceptionV3~\cite{inception} and DenseNet121~\cite{densenet} on CUB dataset. We can find that our method surpasses baseline (SnapMix) with a considerable margin both with two backbones. This demonstrates the effectiveness of our method not only on ResNet but also on other convolution architectures.

\begin{table}[ht]
\caption{Performance comparison(Acc.\%) of using other backbones on the CUB dataset.}
\centering
\renewcommand\arraystretch{1.5}
\renewcommand\tabcolsep{14.0pt} 
\begin{tabular}{ccc}
\toprule
            & SnapMix & Ours           \\ \hline
InceptionV3 & 85.54    & \textBF{86.65} \\
DenseNet121 & 87.42    & \textBF{87.83} \\ \bottomrule
\end{tabular}
\label{tab:using other backbones}
\end{table}

\begin{table}[ht]
\caption{Effectiveness(Acc.\%) of each component of CEKD compared with original snapmix on CUB dataset.}
\centering
\renewcommand\arraystretch{1.5}
\renewcommand\tabcolsep{8.0pt} 
\begin{tabular}{ccccc}
\toprule
       & SnapMix & only CD & only CE & CEKD  \\ \hline
Res34  & 87.06      & 87.51                   & 87.72                      & \textBF{88.32} \\
Res101 & 88.45      & 88.89                   & 89.34                      & \textBF{89.59} \\ \bottomrule
\end{tabular}
\label{tab:effectiveness of each component}
\end{table}

\begin{table}[!ht]
\caption{Influence of hyperparameters $\lambda_1$ using Res50 as backbone on CUB dataset.}
\centering
\renewcommand\arraystretch{1.5}
\renewcommand\tabcolsep{8.0pt} 
\begin{tabular}{cccccc}
\toprule
$\lambda_{1}$      & 0.1   & 0.3   & 0.5   & 0.7   & 0.9   \\ \hline
Acc(\%) & 88.21 & 88.32 & 88.38 & \textBF{88.41} & 88.35 \\ \bottomrule
\end{tabular}
\label{tab:Influence of hyperparameters}
\end{table}

\subsection{Effectiveness of each component of CEKD}
In order to evaluate the effectiveness of each component of our method, we perform ablation study of each proposed module on CUB dataset. The results are reported in Table.~\ref{tab:effectiveness of each component}. It can be observed that collabrative ensemble (CE) module gain more improvement both with backbone of Res-34 and Res-101. This might because that CE module integrates and optimizes results from all branches while cross distillation (CD) module just conbines two pairs, which significantly improves the robustness of model to noise interference. Each component contributes to the improvement of performance.

\subsection{Influence of hyperparameters}
The parameter $\lambda_{1...6}$ in Eq.~\ref{all loss} plays a pivotal role during model training. For brevity, we only show the results under 
different options for $\lambda_1 \in \{ 0.1,0.3,0.5,0.7,0.9 \}$ and corresponding $\lambda_{2}=1-\lambda_{1}$. Meanwhile, we set $\lambda_{3...6}$ all equal to 0.5 for excluding their impact on $\lambda_1$ and $\lambda_2$. The performance of different $\lambda_{1}$ is shown in Table.~\ref{tab:Influence of hyperparameters}. We can find that the accuracy first increases and then decreases slightly, reaching the highest performance when $\lambda_{1} = 0.7$. The reason may be that the basic performance of teacher network is higher than that of student, thus $L_{KD}^{h_1}$ is more effective on correcting student network than $L_{KD}^{h_2}$ on correcting teacher network.

\begin{figure}[!ht] 
	\centering 
	\includegraphics[width=\linewidth]{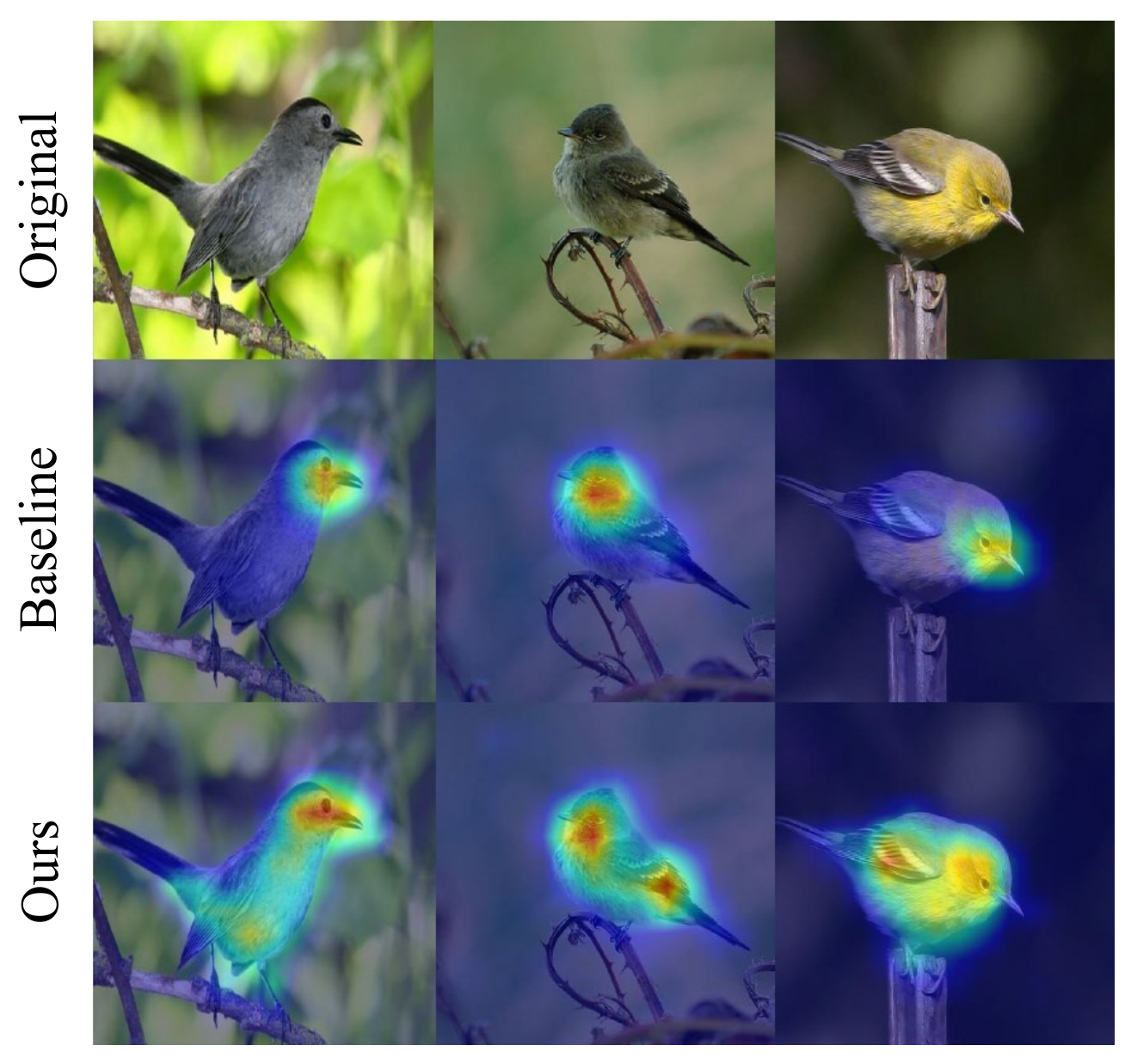} 
	\caption{Visualization of the selecting results from CUB dataset using CAM that compares our method with baseline.} 
	\label{fig:CAM} 
\end{figure}

\subsection{Visualization}
In order to further investigate the effectiveness of the proposed method,  Grad-CAM~\cite{Grad-CAM} is applied  to visualize the images on CUB dataset. Grad-CAM is formulated by weighted summation of feature maps, which marks the regions of interest of the model for classification. As shown in Fig.~\ref{fig:CAM}, the activation maps of our method with baseline are compared. It can be observed observe that baseline only focus on local areas of birds for classification such as the head part, while our method grasps the global feature of birds and focuses more on detail areas (e.g. body parts). This is because we integrates the advantages of different mixed-based methods and complements each other to obtain a more robust model, capturing more subtle and comprehensive fine-grained features.

\section{Conclution}
\label{conclution}
In this paper, we provide a novel insight to combine data augmentation methods with knowledge distillation technique and propose a simple but effective CEKD model for augmenting fine-grained data. Among them, the cross distillation module produces additional supervision to enhance the robustness of the model against noise, and the collaborative ensemble module ensembles and optimizes all final output logits to alleviate the target conflict problem. We have conducted extensive experiments on three fine-grained benchmarks, and the proposed model achieves state-of-the-art performance. With the backbone of ResNet-101, CEKD obtains the accuracy of 89.59\%, 95.96\% and 94.56\% in three datasets respectively, outperforming state-of-the-art API-Net  by 0.99\%, 1.06\% and 1.16\%. Moreover, we provide an explanation on higher accuracy teacher network achieves when MixUp is adopted as student network compared with CutMix. For the sake of brevity, our experiments only evaluated three mix-based data augmentation methods, and the experimental results will be limited. It is an unsolved problem of how to try any combination of data augmentation methods, such as cutout, random clip, random rotation and so on. Further work might also explore how to extend our approach to more general situations, such as compatibility with an arbitrary finite number of branches.

\section*{Acknowledgements}
This work was supported by a Grant from National Natural Science Foundation of China 61972121, Zhejiang provincial Natural Science Foundation of China LY21F020015, Science and Technology Program of Zhejiang Province(No.2021C01187).


\end{document}